\documentclass[runningheads]{article}
\usepackage{graphicx}
\graphicspath{ {images/} }
\usepackage[utf8]{inputenc} 
\usepackage[T1]{fontenc}    
\usepackage{hyperref}       
\usepackage{url}            
\usepackage{booktabs}       
\usepackage{amsfonts}       
\usepackage{nicefrac}       
\usepackage{microtype}      
\usepackage{bm}
\usepackage{amsmath}
\usepackage{color}
\usepackage{wrapfig}
\usepackage{algorithm2e}
\usepackage{tcolorbox}
\usepackage{color}
\usepackage{url}
\usepackage{tikz,pgfplots}
\usepackage{pgf}            
\usepackage{multirow}

\usepackage{algorithmic}

\newcommand{\dist}{\mathcal{D}}

\newcommand{\inpts}{{\cal X}}
\newcommand{\obs}{{\bf x}}

\newcommand{\view}{v}
\newcommand{\obsel}[1]{x^{#1}}
\newcommand{\obsv}{\obsel{\view}}
\newcommand{\lbl}{\mathbf{y}}
\newcommand{\R}{\mathbb{R}}
\renewcommand{\S}{\mathcal{S}}
\newcommand{\minib}{\mathcal{B}}
\newcommand{\EN}{\texttt{EN}}
\newcommand{\FR}{\texttt{FR}}
\newcommand{\GR}{\texttt{GR}}
\newcommand{\IT}{\texttt{IT}}
\newcommand{\SP}{\texttt{SP}}
\renewcommand{\c}{\mathbf{c}}
\newcommand{\MKL}{\texttt{MKL}}
\newcommand{\cocla}{\texttt{co-classif}}
\newcommand{\mvb}{\texttt{mv}$_b$}

\newcommand{\ToneGtwo}{{\scriptsize\textbf{T}$_{\EN \tilde{v}}$}\normalsize}
\newcommand{\TtwoGone}{{\scriptsize\textbf{T}$_{\tilde{\EN} v}$}\normalsize}
\usepackage{mathtools}

\DeclarePairedDelimiterX{\infdivx}[2]{(}{)}{%
  #1\;\delimsize\|\;#2%
}
\newcommand{\infdiv}{\infdivx}

\newcommand{\noview}{\perp}
\newcommand{\lbls}{{\cal Y}}

\newcommand{\BiGame}{Cond$^2$GANs}
\newcommand{\TBiGame}{\texttt{\BiGame}}

\newtheorem{theo}{Theorem}
\newtheorem{assump}{Assumption}
\newtheorem {coro}{Corollary}
\newtheorem {prop}{Proposition}

\newcommand{\CQui}{\texttt{C15}}
\newcommand{\CCAT}{\texttt{CCAT}}
\newcommand{\ECAT}{\texttt{ECAT}}
\newcommand{\EVin}{\texttt{E21}}
\newcommand{\GCAT}{\texttt{GCAT}}
\newcommand{\MOnz}{\texttt{M11}}
\begin{document}
\title{Biconditional Generative Adversarial Networks for Multiview Learning with Missing Views}
\author{
\begin{tabular}{cc}
\multicolumn{2}{c}{Anastasiia Doinychko$^{1,2}$, Massih-Reza Amini$^1$}\\
$^1$Université Grenoble Alpes & $^2$Mentor Graphics \\
LIG/IMAG, 700 av. centrale & 110 Rue Blaise Pascal\\
38401 Saint-Martin d'Hères & 38330 Montbonnot-Saint-Martin\\
France & France\\
\multicolumn{2}{c}{\url{FirstName.Lastname@univ-grenoble-alpes.fr}}
\end{tabular}
}
\date{}
\maketitle              
\begin{abstract}
In this paper, we present a conditional GAN with two generators and a common discriminator for multiview learning problems where observations have two views, but one of them may be missing for some of the training samples. This is for example the case for multilingual collections where documents are not available in all languages. Some studies tackled this problem by assuming the existence of view generation functions to approximately complete the missing views; for example Machine Translation to translate documents into the missing languages. These functions generally require an external resource to be set and their quality has a direct impact on the performance of the learned multiview classifier over the completed training set. Our proposed approach addresses this problem by jointly learning the missing views and the multiview classifier using a tripartite game with two generators and a discriminator. Each of the generators is associated to one of the views and tries to fool the discriminator by generating the other missing view conditionally on the corresponding observed view. The discriminator then tries to identify if for an observation, one of its views is completed by one of the generators or if both views are completed along with its class. Our results on a subset of Reuters RCV1/RCV2 collections show that the discriminator achieves significant classification performance; and that the generators learn the missing views with high quality without the need of any consequent external resource.

\end{abstract}
\section{Introduction}
We address the problem of multiview learning with Generative Adversarial Networks (GANs) in the case where some observations may have missing views without there being an external resource to complete them. This is a typical situation in many applications where different sources generate different views of samples unevenly; like text information present in all Wikipedia pages while images are more scarce. 
Another example is multilingual text classification where documents are available in two languages and share the same set of classes while some are just written in one language. Previous works supposed the existence of view generating functions to complete the missing views before deploying a learning strategy \cite{AminiEtAlNIPS22}. However, the performance of the global multiview approach is biased by the quality of the generating functions which generally require external resources to be set. The challenge is hence to learn an efficient model from the multiple views of training data without relying on an extrinsic approach to generate altered views for samples that have missing ones. 

In this direction, GANs provide a propitious and broad approach with a high ability to seize the underlying distribution of the data and create new samples \cite{GoodfellowEtAlNIPS27}. These models have been mostly applied to image analysis and major advances have been made on generating realistic images with low variability \cite{DentonEtAlNIPS28,OdenaEtAlICML17,RadfordICLR16}. GANs take their origin from the game theory and are formulated as a two players game formed by a generator $G$ and a discriminator $D$. The generator takes a noise $z$ and produces a sample $G(z)$ in the input space, on the other hand the discriminator determines whenever a sample comes from the true distribution of the data or if it is generated by $G$. Other works included an inverse mapping from the input to the latent representation, mostly referred to as BiGANs, and showed the usefulness of the learned feature representation for auxiliary discriminant problems \cite{DonahueEtAlICLR17,DumoulinEtAlICLR17}. This idea paved the way for the design of efficient approaches for generating coherent synthetic views of an input image \cite{YuEtAlijcai2018,MaEtAlNIPS30,ChenD16a}.

In this work, we propose a GAN based model for bilingual text classification, called {\TBiGame}, where some training documents are just written in one language. The model learns the representation of missing versions of bilingual documents jointly with the association to their respective classes, and is composed of a discriminator $D$ and two generators $G_1$ and $G_2$ formulated as a tripartite game. For a given document with a missing version in one language, the corresponding generator induces the latter conditionally on the observed one. The training of the generators is carried out by minimizing a regularized version of the cross-entropy measure proposed for multi-class classification with GANs \cite{SpringenbergICLR16} in a way to force the models to generate views such that the completed bilingual documents will have high class assignments. At the same time, the discriminator learns the association between documents and their classes and distinguishes between observations that have their both views and those that got a completed view by one of the generators. This is achieved by minimizing an aggregated cross-entropy measure in a way to force the discriminator to be certain of the class of observations with their complete views and uncertain of the class of documents for which one of the versions was completed. The regularization term in the objectives of generators is derived from an adapted feature matching technique \cite{SalimansEtAlNIPS29} which is an effective way for preventing from situations where the models become unstable; and which leads to fast convergence. 

We demonstrate that generated views allow to achieve state-of-the-art results on a subset of Reuters RCV1/RCV2 collections compared to multiview approaches that rely on Machine Translation (MT) for translating documents into languages in which their versions do not exist; before training the models. Importantly, we exhibit qualitatively that generated documents have meaningful translated words bearing similar ideas compared to the original ones; and that, without employing any large external parallel corpora to learn the translations as it would be the case if MT were used. More precisely, this work is the first to~:
\begin{itemize}
    \item Propose a new tripartite GAN model that makes class prediction along with the generation of high quality document representations in different input spaces in the case where the corresponding versions are not observed (Section \ref{sec:Model}); 
    \item Achieve state-of-the art performance compared to multiview approaches that rely on external view generating functions on multilingual document classification; and which is  another challenging application than image analysis which is the domain of choice for the design of new GAN models (Section \ref{sec:results});
    \item Demonstrate the value of the generated views within our approach compared to when they are generated using MT (Section \ref{sec:results}).
\end{itemize}
\section{Related work}

Multiview learning has been an active domain of research these past few years. Many advances have been made on both theoretic and algorithmic sides \cite{Blum:1998,GoyalElAlECML17}. The three main families of techniques for (semi-)supervised learning are (kernel) Canonical Correlation Analysis (CCA), Multiple kernel learning (MKL) and co-regularization. CCA finds pairs of highly correlated subspaces between the views that is used for mapping the data before training, or integrated in the learning objective \cite{BachetAl02,FarquharetAl06}. MKL  considers one kernel per view and different approaches have been proposed for their learning.   In one of the earliest work, \cite{BachEtal04} proposed an efficient algorithm based on sequential minimization techniques for learning a corresponding support vector machine defined over a convex nonsmooth optimization problem. Co-regularization techniques tend to minimize the disagreement between the single-view classifiers over their outputs on unlabeled examples by adding a regularization term to the objective function \cite{SindhwaniEtAl08}. Some approaches have also tackled the tedious question of combining the predictions of the view specific classifiers \cite{tianElAlAISTATS19}. However all these techniques assume that  views of a sample are complete and available during training and testing. 

Recently, many other studies have considered the generation of multiple views from a single input image using GANs \cite{MaEtAlNIPS30,YuEtAlijcai2018,ZhaoEtAlmm2018} and have demonstrated the intriguing capacity of these models to generate coherent unseen views. The former approaches rely mostly on an encoder-encoder network to first map images into a latent space and then generate their views using an inverse mapping. This is a very exciting problem, however, our learning objective differs from these approaches as we are mostly interested in the classification of muti-view samples with missing views. The most similar work to ours that uses GANs for multiview classification is probably \cite{ChenD16a}. This approach generates missing views of images in the same latent space than the input image, while {\TBiGame} learns the representations of the missing views in their respective  input spaces conditionally on the observed ones which in general are from other feature spaces. Furthermore, {\TBiGame} benefits from low complexity and stable convergence which has been shown to be lacking in the previous approach. 

Another work which has considered multiview learning with incomplete views, for also document classification, is \cite{AminiEtAlNIPS22}. The authors proposed a Rademacher complexity bounds for a multiview Gibbs classifier trained on multilingual collections where the missing versions of documents have been generated by Machine Translation systems. Their bounds exhibit a term corresponding to the quality of the MT system  generating the views.  The bottleneck is that MT systems depend on external resources, and they require a huge amount of parallel collections containing documents and their translations in all languages of interest for their tuning.  For rare languages, this can ultimately affect the performance of the learning models. Regarding these aspects our proposed approach differs from all the previous studies, as we do not suppose the existence of parallel training sets nor MT systems to generate the missing versions of the training observations.
\section{\BiGame}
In the following sections, we first present the basic definitions which will serve to our problem setting, and then the proposed model for multiview classification with missing views. 
\subsection{Framework and problem setting}

We consider multiclass classification problems, where a bilingual document is defined as a sequence $\obs = (\obsel{1},\obsel{2})\in\inpts$ that belongs to one and only one class $\lbl\in \lbls = \{0,1\}^K$. The class membership indicator vector $\lbl=(y_k)_{1\le k\le K}$, of each bilingual document, has all its components equal to $0$ except the one that indicates the class associated with the example which is equal to one. Here we suppose that $\inpts= \left(\inpts_1 \cup\{\noview\}\right)\times\left(\inpts_{2} \cup\{\noview\}\right)$, where $\obsv = \noview$ means that the $v$-th view is not observed. Hence, each observed view $\obsv\in\obs$ is such that $\obsv\neq \noview$ and it provides a representation of $\obs$ in a corresponding input space $\inpts_v\subseteq \R^{d_v}$. Following the conclusions of the co-training study \cite{Blum:1998}, our framework is based on the following main assumption~:
\begin{assump}[\cite{Blum:1998}]
\label{assump:informativness}
Observed views are not completely correlated, and are equally informative.
\end{assump}
Furthermore, we assume that each example $(\obs,\lbl)$ is identically and independently distributed (i.i.d.) according to a fixed yet unknown distribution $\dist$ over $\inpts\times \lbls$, and that at least one of its views is observed. Additionally, we suppose to have access to a training set $\S=\{(\obs_i,\lbl_i); i\in\{1,\ldots,m\}\}=\S_F\sqcup \S_1\sqcup \S_2$ of size $m$ drawn i.i.d. according to $\dist$, where $\S_F=\{\left(\left(\obsel{1}_i,\obsel{2}_i\right),\lbl_i\right)\mid i\in\{1,\ldots,m_F\}\}$ denotes the subset of training samples with their both complete views and $\S_1=\{\left(\left(\obsel{1}_i,\noview\right),\lbl_i\right)\mid i\in\{1,\ldots,m_1\}\}$ (respectively $\S_2=\{\left(\left(\noview, \obsel{2}_i\right),\lbl_i\right)\mid i\in\{1,\ldots,m_2\}\}$) is the subset of training samples with their second (respectively first) view that is not observed (i.e. $m=m_F+m_1+m_2$). 

It is possible to address this problem using existing techniques; for example, by learning singleview classifiers independently on the examples of $\S\sqcup\S_1$ (respectively $\S\sqcup\S_2$) for the first (respectively second) view. To make prediction, one can then combine the outputs of  the classifiers \cite{tianElAlAISTATS19} if both views of a test example are observed; or otherwise, use one of the outputs corresponding to the observed view. Another solution is to apply multiview approaches over the training samples of $\S_F$; or over the whole training set $\S$ by completing the views of examples in $\S_1$ and $\S_2$ beforhand using external view generation functions.

\subsection{The Tripartite Game}
\label{sec:Model}

As an alternative, the learning objective of our proposed approach is to generate the missing views of examples in $\S_1$ and $\S_2$, jointly with the learning of the association between the multiview samples (with all their views complete or completed) and their classes. The proposed model consists of three \begin{wrapfigure}[22]{r}{0.67\textwidth}
\includegraphics[width=0.65\textwidth]{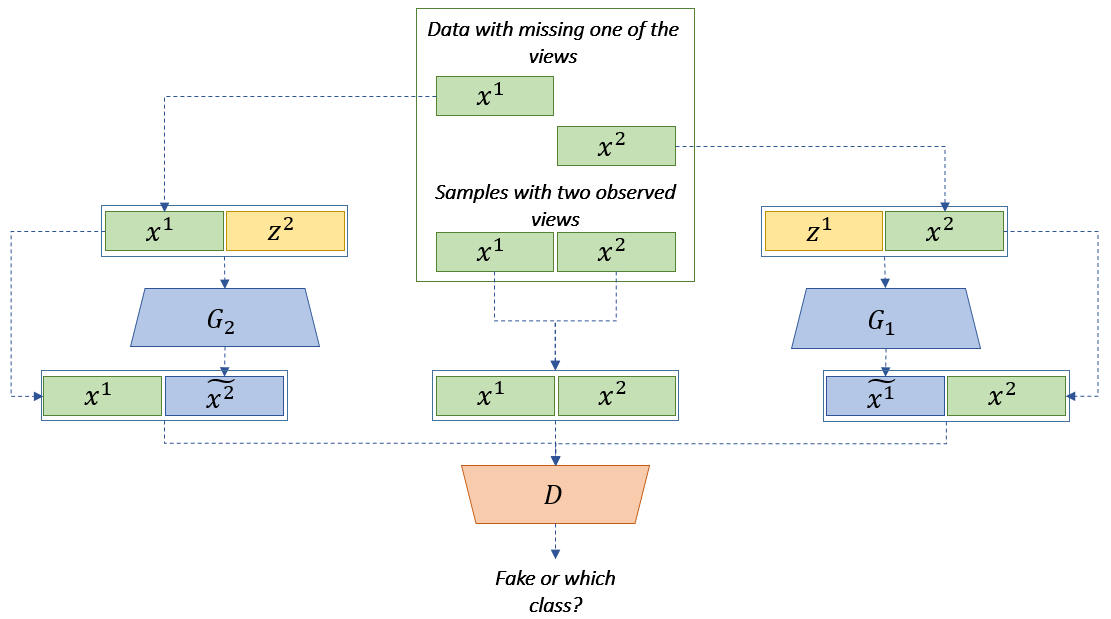}
\centering
\caption{A visual representation of the proposed GAN model composed of three neural networks; a discriminator $D$ and two generators $G_1$ and $G_2$. The missing view of an observation is completed by the corresponding generator conditionally on its observed view. The discriminator is trained to recognize between observations having their views completed and those with complete initial views as well as their classes.}
\label{fig:ourmodel}
\end{wrapfigure} neural networks that are trained using an objective implementing a three players game between a discriminator, $D$, and two generators, $G_1$ and $G_2$. The game that these models play is depicted in Figure \ref{fig:ourmodel} and it can be summarized as follows. At each step, if an observation is chosen with a missing view, the corresponding generator -- $G_1$ (respectively $G_2$) if the first (respectively second) view is missing -- produces the view from random noise conditionally on the observed view in a way to fool the discriminator. On the other hand, the discriminator takes as input an observation with both of its views complete or completed and, classifies it if the views are initially observed or tells if a view was produced by one of the generators. Formally, both generators $G_1$ and $G_2$ take as input; samples from respectively the training subsets $\S_2$ and $\S_1$; as well as random noise drawn from uniform distribution defined over the input space of the missing view and produce the corresponding  pseudo-view, which is missing; i.e. $G_1(z^1,x^2)=\tilde{x}^1$ and $G_2(x^1, z^2)=\tilde{x}^2$. These models are learned in a way to replicate the conditional distributions $p(x^1|x^2,z^1)$ and $p(x^2|x^1,z^2)$; and inherently define two probability distributions, denoted respectively by $p_{G_1}$ and $p_{G_2}$, as the distribution of samples if both views where observed i.e.  $(\tilde{x}^1,x^2)\sim p_{G_1}(x^1,x^2)$, $(x^1,\tilde{x}^2)\sim p_{G_2}(x^1,x^2)$. On the other hand, the discriminator takes as input a training sample; either from the set $\S_F$, or from one of the training subsets $\S_1$ or $\S_2$ where the missing view of the example is generated by one of the generators accordingly.  The task of $D$ is then to recognize observations from $\S_1$ and $\S_2$ that have completed views by $G_1$ and $G_2$ and to classify examples from $\S$ to their true classes. To achieve this goal we add a fake class, $K+1$, to the set of classes, $\lbls$, corresponding to samples that have one of their views generated by $G_1$ or $G_2$. The dimension of the discriminator's output is hence set to $K+1$ which by applying softmax is supposed to estimate the posterior probability of classes for each multiview observation (with complete or completed views) given in input. For an observation $\obs\in\inpts$, we use $D_{K+1}(\obs)= p_D(y=K+1 | \obs)$ to estimate the probability that one of its views is generated by $G_1$ or $G_2$. As the task of the generators is to produce good quality views such that the observation with the completed view will be assigned to its true class with high probability, we follow \cite{SalimansEtAlNIPS29} by supplying the discriminator to not get fooled easily as stated in the following assumption~:
\begin{assump}[\cite{SalimansEtAlNIPS29}]
\label{axiom}
An observation $\obs$ has one of its views generated by $G_1$ or $G_2$; if and only if $D_{K+1}(\obs) > \sum_{k=1}^K D_k(\obs)$.
\end{assump}
In the case where; $D_{K+1}(\obs)\le \sum_{k=1}^K D_k(\obs)$ the observation $\obs$ is supposed to have its both views observed and it is affected to one of the classes following the rule; $\mathop{\max}_{k=\{1,\ldots,K\}} D_k(\obs)$.
The overall learning objective of {\TBiGame} is to train the generators to produce realistic views indistinguishable with the real ones, while the discriminator is trained to classify multiview observations having their complete views and to identify view generated samples. If we denote by $p_{real}$ the marginal distribution of multiview observations with their both views observed (i.e. $(x^1,x^2)=p_{real}(x^1,x^2)$); the above procedure resumes to the following tripartite minmax game with value function $V(D,G_1,G_2)$~:
\begin{align} 
\label{eq:V}
    \mathop{\max}_{D} \mathop{\min}_{G_1,G_2} V(D,G_1,G_2) & = ~\mathbb{E}_{(x^1,x^2)\sim p_{real}}\left[\log p_D(y<K+1 |x^1,x^2)\right] \nonumber \\ &+ \frac{1}{2} \mathbb{E}_{(\tilde{x}^1, x^2)\sim p_{G_1}}\left[\log p_D(y=K+1 | \tilde{x}^1, x^2)\right] \\ &+ \frac{1}{2}\mathbb{E}_{(x^1, \tilde{x}^2)\sim p_{G_2}}\left[\log p_D(y=K+1 | x^1, \tilde{x}^2)\right] \nonumber
\end{align}
Note that, following Assumption \ref{assump:informativness}, we impose the generators to produce equally informative views by assigning the same weight  to their corresponding terms in $V$ (Eq. \ref{eq:V}).

\subsection{Analyses and convergence}

From the outputs of the discriminator we build an auxiliary function $\mathbf{D}$ equal to the sum of the first $K$ outputs associated to the true classes~:
\begin{equation}
\label{eq:Auxilliary}
\forall \obs\in\inpts; \mathbf{D}(\obs)=\sum_{k=1}^K p_D(y=k\mid \obs)
\end{equation}
In this following, we provide a theoretical analysis of {\TBiGame} involving the auxiliary function $\mathbf{D}$ (Eq. \ref{eq:Auxilliary}) under nonparametric hypotheses. 
\begin{prop}
For fixed generators $G_1$ and $G_2$, the minmax game defined in (Eq. \ref{eq:V}) leads to the following optimal discriminator $\mathbf{D}^*_{G_1,G_2}$~:
\begin{equation}
    \label{eq:D*}
    \mathbf{D}_{G_1,G_2}^*(x^1,x^2)=\frac{p_{real}(x^1,x^2)}{p_{real}(x^1,x^2)+p_{G_{1,2}}(x^1,x^2)},
\end{equation}
where $p_{G_{1,2}}(x^1,x^2)=\frac{1}{2}(p_{G_1}(x^1,x^2)+p_{G_2}(x^1,x^2))$.
\end{prop}
\textbf{Proof.} The proof follows from \cite{GoodfellowEtAlNIPS27}. Let 
\[
\forall \obs=(x^1,x^2), \mathbf{D}(\obs)=\sum_{k=1}^K D_k(\obs)
\]
From Assumption \ref{axiom}, and the fact that for any observation $\obs$ the outputs of the discriminator sum to one i.e. $ \sum_{k=1}^{K+1}D_{k}(\obs)=1$, the value function $V$ writes~:
{\small
\begin{multline*}
    V(\mathbf{D},G_1,G_2)=\iint \log( \mathbf{D}(x^1,x^2))p_{real}(x^1,x^2)dx^1dx^2+\\
    \frac{1}{2}\iint \log(1-\mathbf{D}(x^1,x^2))p_{G_1}(x^1,x^2)dx^1dx^2+\frac{1}{2}\iint \log(1-\mathbf{D}(x^1,x^2))p_{G_2}(x^1,x^2)dx^1dx^2
\end{multline*}}
For any $(\alpha,\beta,\gamma)\in\R^3\backslash \{0,0,0\}$; the function $z\mapsto \alpha \log z+\frac{\beta}{2}\log(1-z)+\frac{\gamma}{2}\log(1-z)$ reaches its maximum at $z=\frac{\alpha}{\alpha+\frac{1}{2}(\beta+\gamma)}$, which ends the proof as the discrimintaor does not need to be defined outside the supports of $p_{data}, p_{G_1}$ and $p_{G_2}$.
$\Box$

By plugging back $\mathbf{D}^*_{G_1,G_2}$ (Eq. \ref{eq:D*}) into the value function $V$ we have the following necessary and sufficient condition for attaining the global minimum of this function~:

\begin{theo}
The global minimum of the function $V(G_1,G_2)$ is attained if and only if 
\begin{equation}
\label{eq:equilibrium1}
p_{real}(x^1,x^2)=\frac{1}{2}(p_{G_1}(x^1,x^2)+p_{G_2}(x^1,x^2)).
\end{equation}
At this point, the minimum is equal to $-\log 4$.
\end{theo}
\textbf{Proof.} By plugging back the expression of $\mathbf{D}^*$ (Eq. \ref{eq:D*}), into the value function $V$, it comes
\small{\begin{multline*}
    V(\mathbf{D}^*,G_1,G_2)=\iint \log\left( \frac{p_{real}(x^1,x^2)}{p_{real}(x^1,x^2)+p_{G_{1,2}}(x^1,x^2)}\right)p_{real}(x^1,x^2)dx^1dx^2+\\
    \iint \log\left(\frac{p_{G_{1,2}}(x^1,x^2)}{p_{real}(x^1,x^2)+p_{G_{1,2}}(x^1,x^2)}\right)p_{G_{1,2}}(x^1,x^2)dx^1dx^2
\end{multline*}}
Which from the definition of the Kullback Leibler (KL) and the Jensen Shannon divergence (JSD) can be rewritten as
\begin{align*}
V(\mathbf{D}^*,G_1,G_2)=&-\log 4+KL\infdiv*{p_{real}}{\frac{p_{real}+p_{G_{1,2}}}{2}}+KL\infdiv*{p_{G_{1,2}}}{\frac{p_{real}+p_{G_{1,2}}}{2}}\\=& -\log 4+2JSD\infdiv*{p_{real}}{p_{G_{1,2}}}
\end{align*}
The JSD is always positive and $JSD\infdiv*{p_{real}}{p_{G_{1,2}}}=0$ if and only if $p_{real}=p_{G_{1,2}}$ which ends the proof $\Box$

From Equation \ref{eq:equilibrium1}, it is straightforward to verify that $p_{real}(x^1,x^2)=p_{G_1}(x^1,x^2)=p_{G_2}(x^1,x^2)$ is a global Nash equilibrium but it may not be unique. In order to ensure the uniqueness, we add the Jensen-Shannon divergence between the distribution $p_{G_1}$ and $p_{real}$ and $p_{G_2}$ and $p_{real}$ the value function $V$ (Eq. \ref{eq:V}) as stated in the corollary below.
\begin{coro}
\label{cor:GNE}
The unique global Nash equilibrium of the augmented value function~:
\begin{equation}
\label{eq:AugmentedV}
\bar{V}(\mathbf{D},G_1,G_2)=V(\mathbf{D},G_1,G_2)+JSD(p_{G_1}||p_{real})+JSD(p_{G_2}||p_{real}),
\end{equation}
is reached if and only if
\begin{equation}
\label{eq:GNE}
p_{real}(x^1,x^2)=p_{G_1}(x^1,x^2)=p_{G_2}(x^1,x^2),
\end{equation}
where $V(\mathbf{D},G_1,G_2)$ is the value function defined in Equation \eqref{eq:V} and $JSD(p_{G_1}||p_{real})$ is the Jensen-Shannon divergence between the distribution $p_{G_1}$ and $p_{real}$.
\end{coro}
\textbf{Proof.}  The proof follows from the positivness of JSD and the necessary and sufficient condition for it to be equal to $0$. Hence, $\bar{V}(\mathbf{D},G_1,G_2)$ reaches it minimum $-\log 4$, iff $p_{G_1}=p_{real}=p_{G_2}$.   $\Box$

This result suggests that at equilibrium, both generators produce views such that observations with their completed view follow the same real distribution than those which have their both views observed. 

\subsection{Algorithm and Implementation}

In order to avoid the collapse of the generators \cite{SalimansEtAlNIPS29}, we perform minibatch discrimination by allowing the discriminator to have access to multiple samples in combination. From this perspective, the minmax game (Eq. \ref{eq:V}) is equivalent to the maximization of a cross-entropy loss, and we use minibatch training to learn the parameters of the three models. The corresponding empirical errors estimated over a minibatch $\minib$ that contains $m_b$ samples from each of the sets $\S_F$, $\S_1$ and $\S_2$ are~:
{\small  \begin{align}
\mathcal{L}_D(\minib)& =  -\frac{1}{m_b}\!\!\sum_{\obs\in\minib\cap\S_F}\frac{1}{K+1}\sum_{k=1}^{K}y_k\log\big[D_k(x^1, x^2)\big]\\ & - \frac{1}{2m_b}\!\!\sum_{\obs\in\minib\cap\S_1}\!\!\log\big[D_{K+1}(G_1(z^1,x^2), x^2))\big]    -\frac{1}{2m_b}\!\!\sum_{\obs\in\minib\cap\S_2}\!\!\log\big[D_{K+1}(x^1, G_2(x^1,z^2))\big] \nonumber\\
    \mathcal{L}_{G_v}(\minib) & =  -\frac{1}{m_b}\sum_{\obs\in\minib\cap\S_v}\frac{1}{K+1}\sum_{k=1}^{K}y_k\log\big[D_k(G_v(z^v, x^{3-v}), x^{3-v})\big]  +\mathcal{L}_{FM}^v; v\in\{1,2\} \label{eq:LG1}
\end{align}
}

 
\begin{wrapfigure}[16]{r}{0.65\textwidth}
\vspace{-4mm} \hrule
Minibatch stochastic training of \TBiGame
\hrule
\vspace{3mm}\textbf{Input:} A training set $\S=\S_F\sqcup\S_1\sqcup\S_2$\\
\textbf{Initialization:} 
Size of minibatches, $m_b$\\
Use $Xavier$ initializer to initialize discriminator and  generators parameters, respectively $\theta^{(0)}_d,\theta^{(0)}_{g_1}, \theta^{(0)}_{g_2}$\\
\For{$i = 0\ldots T-1$}{
Sample randomly a minibatch $\minib_i$ of size $3m_b$ from $\S_1$, $\S_2$ and $\S_F$; create minibatches of noise vector $z^1, z^2$ from $\mathcal{U}(-1,1)$\;\\
$\theta_d^{(i+1)} \leftarrow  Adam(\mathcal{L}_D(\minib_i), \theta_d^{(i)},\alpha,\beta)$\; {\color{gray}{\scriptsize \# Update of $D$}}\\
$\theta_{g_1}^{(i+1)} \leftarrow Adam(\mathcal{L}_{G_1}(\minib_i), \theta_{g_1}^{(i)},\alpha,\beta)$\; {\color{gray}{\scriptsize \# Update of $G_1$}}\\
$\theta_{g_2}^{(i+1)} \leftarrow Adam(\mathcal{L}_{G_2}(\minib_i), \theta_{g_2}^{(i)},\alpha,\beta)$\; {\color{gray}{\scriptsize \# Update of $G_2$}}\\
}
\hrule
\end{wrapfigure}

In order, to be inline with the premises of Corollary \ref{cor:GNE}; we empirically tested different solutions and the most effective one that we found was the feature matching technique  proposed in \cite{SalimansEtAlNIPS29}, which addressed the problem of instability for the learning of generators by adding a penalty term $\mathcal{L}_{FM}^v=\|\mathbb{E}_{p_{real}}f(x^1,x^2) - \mathbb{E}_{p_{G_v}}f(x^{3-v},G_v(x^v))\|$ $,  v\in\{1,2\}$ to their corresponding objectives (Eq. \eqref{eq:LG1}). Where, $\|.\|$ is the $\ell_2$ norm and $f$ is the sigmoid activation function on an intermediate layer of the discriminator. The overall algorithm of {\TBiGame} is shown above. The parameters of the three neural networks are first initialized using {\it Xavier}. For a given number of iterations $T$, minibatches of size $3m_b$ are randomly sampled from the sets $\S_F$, $\S_1$ and $\S_2$. Minibatches of noise vectors are randomly drawn from the uniform distribution. Models parameters of the discriminator and both generators are then sequentially updated using \textit{Adam} optimization algorithm \cite{kingmaEtAlICLR2015}. We implemented our method by having two layers neural networks for each of the components of \TBiGame. These neural nets are composed of 200 nodes in hidden layers with a sigmoid activation function. Since the values of the generated samples are supposed to approximate any possible real value, we do not use the activation function in the outputs of both generators.\footnote{We will release the code for reproducibility and research purpose.}

\section{Experiments}

In this Section, we present experimental results aimed at evaluating how the generation of views by {\TBiGame} can help to take advantage of existing training examples, with many having an incomplete view, in order to learn an efficient classification function. We perform experiments on a publicly available collection, extracted from Reuters RCV1/RCV2, that is proposed for multilingual multiclass text categorization\footnote{\url{https://archive.ics.uci.edu/ml/datasets/Reuters+RCV1+RCV2+Multilingual,+Multiview+Text+Categorization+Test+collection}} (Table \ref{table:ReutersDataSet}). The dataset contains feature vectors of documents originally presented in five languages (\EN, \FR, \GR, \IT, \SP). In our experiments, we consider four pairs of languages with always English as one of the views ((\EN,\FR),(\EN,\SP),(\EN,\IT),(\EN,\GR)). Documents in different languages belong to one and only one class within the same set of classes ($K=6$); and they also have translations into all the other languages. These translations are obtained from a state-of-the-art Statistical Machine Translation system \cite{Ueffing:2007} trained over the  Europal parallel collection using about $8.10^6$ sentences for the $4$ considered pairs of languages.\footnote{\url{http://www.statmt.org/europarl/}}

\begin{table}[h]
\caption{The statistics of RCV1/RCV2 Reuters data collection used in our experiments.}
\label{table:ReutersDataSet}
\begin{center}
\hfill
\vspace{-9mm}\begin{tabular}{c c c | c}
\hline
Language     &  \# docs  &   (\%)   &  dim \\\hline
\EN         &  $18,758$ &  $16.78$  &  $21,531$ \\
\FR         &  $26,648$ &  $23.45$  &  $24,893$ \\
\GR         &  $29,953$ &  $26.80$  &  $34,279$ \\
\IT         &  $24,039$ &  $21.51$  &  $15,506$ \\
\SP         &  $12,342$ &  $11.46$  &  $11,547$ \\ \cline{1-2}
Total        & $111,740$ &           & \\
\hline
\end{tabular}
\hfill
\begin{tabular}{c c c}
\hline
Class        &      Size (all lang.)    & (\%) \\\hline
\CQui        &     $18,816$  &   $16.84$   \\
\CCAT        &     $21,426$  &   $19.17$   \\
\EVin        &     $13,701$  &   $12.26$   \\
\ECAT        &     $19,198$  &   $17.18$   \\
\GCAT        &     $19,178$  &   $17.16$   \\
\MOnz        &     $19,421$  &   $17.39$  \\
\hline \end{tabular}
\hfill~
\end{center}
\end{table}

\subsection{Experimental Setup}

 In our experiments, we consider the case where the number of training documents having their two versions is much smaller than those with only one of their available versions (i.e. $m_F\ll m_1+m_2$). This corresponds to the case where the effort of gathering documents in different languages is much less than translating them from one language to another.  To this end, we randomly select $m_F=300$ samples having their both views, $m_1 = m_2 = 6000$ samples with one of their views missing and the remaining samples without their translations for test. In order to evaluate the quality of generated views by {\TBiGame} we considered two scenarios. In the first one (denoted by \ToneGtwo), we test on English documents by considering the generation of these documents with respect to the other view ($v\in\{\FR,\GR,\IT,\SP\}$) using the corresponding generator. In the second scenario (denoted by \TtwoGone), we test on documents that are written in another language than English by considering their generation on English provided by the other generator.  For evaluation, we test the following four classification approaches along with {\TBiGame}; one singleview approach and four multiview approaches. In the singleview approach (denoted by {$\c_v$}) classifiers are the same than the discriminator and they are trained on the part of the training set with examples having their corresponding view observed. The multiview  approaches are {\MKL} \cite{BachEtal04}, co-classification ({\cocla}) \cite{AminiEtAlMLJ10}, unanimous vote ( {\mvb}) \cite{AminiEtAlNIPS22}. Results are evaluated over the test set using the accuracy and the F$_1$ measure which is the harmonic average of precision and recall. The reported performance are averaged over 20 random (train/test) sets, and the parameters of Adam optimization algorithm are set to $\alpha = 10^{-4}$, $\beta = 0.5$. 

\subsection{Experimental Results}
\label{sec:results}

\paragraph{On the value of the generated views.}
We start our evaluation by comparing the $F_1$ scores over the test set, obtained with {\TBiGame} and a neural network having the same architecture than the discriminator $D$ of {\TBiGame}  trained over the concatenated views of documents in the training set where the missing views are generated by Machine Translation. Figure \ref{fig:Comparison} shows these results.  Each point represents a class, where its abscissa (resp. ordinate) represents  the test F$_1$ score of the Neural Network trained using MT (resp. one of the generators of {\TBiGame}) to complete the missing views.  All of the classes, in the different language pair scenarios, are above the line of equality, suggesting that the generated views by {\TBiGame} provide 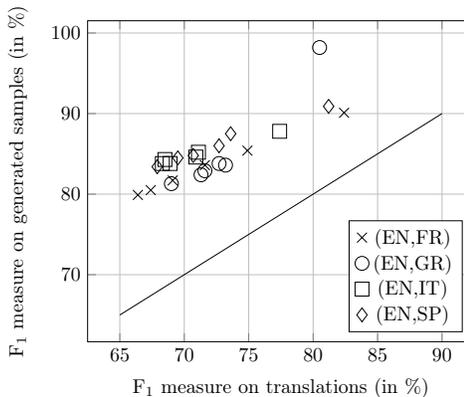
\begin{wrapfigure}[19]{r}{0.64\textwidth}
\vspace{-2mm}  \centering
  \begin{tikzpicture}[scale=0.75]
	\begin{axis}[
		xlabel=F$_1$ measure on translations (in $\%$),
		ylabel=F$_1$ measure on generated samples (in $\%$),
		grid=major,
		legend pos=south east
]
  	\addlegendentry{(EN,FR)}
  	\addplot[only marks,color=black,mark=x, mark size=3.5pt] coordinates {
		(82.4, 90.1)
		(74.9, 85.4)
		(67.4, 80.5)
		(66.4, 79.9)
		(69.1, 81.7)
		(71.6, 83.6)
	};
	\addlegendentry{(EN,GR)}
  	\addplot[only marks,color=black,mark=o, mark size=3.5pt] coordinates {
		(80.5, 98.2)
		(72.7, 83.8)
		(71.6, 82.9)
		(69, 81.3)
		(71.3, 82.4)
		(73.2, 83.6)
	};
	\addlegendentry{(EN,IT)}
  	\addplot[only marks,color=black,mark=square, mark size=3.5pt] coordinates {
		(77.4, 87.8)
		(70.9, 84.6)
		(68.5, 84.3)
		(68.3, 83.8)
		(68.9, 83.8)
		(71.1, 85.2)
	};
	\addlegendentry{(EN,SP)}
  	\addplot[only marks,color=black,mark=diamond,mark size=3.5pt] coordinates {
		(81.2, 90.9)
		(73.6, 87.5)
		(69.5, 84.5)
		(67.9, 83.4)
		(70.7, 84.8)
		(72.7, 86)
	};
	\addplot[ultra thin, domain=65:90]{x};
	\end{axis}
	  \end{tikzpicture}
	  
  \caption{F$_1$-score per class measured for test predictions made by a Neural-Network, with the same architecture than the discriminator of {\TBiGame}, and trained over documents where their missing views are generated by MT, or by $G_1$ or $G_2$. }
\label{fig:Comparison}
\end{wrapfigure}higher value information than translations provided by MT for learning the Neural Network. This is an  impressive finding, as the resources necessary for the training of MT is large ($8.10^6$ pairs of sentences and their translations); while {\TBiGame} does both view completion and discrimination using only the available training data. This is mainly because  both generators induce missing views with the same distribution than real pairs of views as stated in Corollary \ref{cor:GNE}.

\paragraph{Comparison between multiview approaches.} We now examine the gains, in terms of accuracy, of learning the different multiview approaches on a collection where for other approaches than {\TBiGame} the missing views are completed by one of the generators of our model.   Table \ref{tab:AccGenviews-1} summarizes these results obtained by {\TBiGame}, \MKL, \cocla, and \mvb{} for both test scenarios. In all cases \TBiGame{}, provides significantly better results than other approaches. This provides empirical evidence of the effectiveness of  the joint  view generation and class prediction of {\TBiGame{}}. Furthermore, {\MKL}, {\cocla} and {\TBiGame{}} are binary classification models and tackle the multiclass classification case with one vs all strategy making them to suffer from class imbalance problem. Results obtained using the F$_1$ measure are in line with those of Table \ref{tab:AccGenviews-1} and they are not reported for the sake of space.

\begin{figure}[h!]
  \centering
  \begin{tabular}{ccc}
  \begin{tikzpicture}[scale=0.5]
	\begin{axis}[
	    xmode=log,
        log ticks with fixed point,
		xlabel=$\#$ training samples with their observed \textbf{IT} view,
		ylabel=F$_1$ (in $\%$),
		grid=major,
		legend pos=south east
]
  	\addlegendentry{\TBiGame}
  	\addplot[color=black,mark=*] coordinates {
		(25,72.9)
		(100,77.9)
		(500,82.7)
		(1000,84.6)
	};
  	\addlegendentry{singleview classifier}
	\addplot[color=gray,mark=o] coordinates {
		(25, 69.1)
		(100, 75.8)
		(500, 80.6)
		(1000, 83.4)
	};
	\end{axis}
	  \end{tikzpicture}
&
  \begin{tikzpicture}[scale=0.5]
	\begin{axis}[
	    xmode=log,
        log ticks with fixed point,
		xlabel=$\#$ training samples with their observed \textbf{SP} view,
		grid=major,
		legend pos=south east
]
  	\addlegendentry{\TBiGame}
  	\addplot[color=black,mark=*] coordinates {
		(25, 81.6)
		(100, 84.7)
		(500, 86.8)
		(1000, 88.7)
	};
  	\addlegendentry{singleview classifier}
	\addplot[color=gray,mark=o] coordinates {
		(25, 80.2)
		(100, 83.9)
		(500, 86.2)
		(1000, 88.2)
	};
	\end{axis}
	  \end{tikzpicture} &
  \begin{tikzpicture}[scale=0.5]
	\begin{axis}[
	    xmode=log,
        log ticks with fixed point,
		xlabel=$\#$ training samples with their observed \textbf{EN} view,
		grid=major,
		legend pos=south east
]
  	\addlegendentry{\TBiGame}
  	\addplot[color=black,mark=*] coordinates {
		(25, 74.8)
		(100, 80)
		(500, 84.9)
		(1000, 86.7)
	};
  	\addlegendentry{singleview classifier}
	\addplot[color=gray,mark=o] coordinates {
		(25, 72.7)
		(100, 78.9)
		(500, 84.4)
		(1000, 86.1)
	};
	\end{axis}
	  \end{tikzpicture}\\
	
(a) $m_{\EN}=6000$ & (b) $m_{\EN}=6000$ & (c) $m_{\GR}=6000$ 

\end{tabular}

  \caption{F$_1$ measure of {\TBiGame} and a singleview  classifier ($\c_v$) for an increasing number of training samples with the corresponding view that is observed. The number of training examples corresponding to the other view ($m_{v\!\!\!\backslash} =6000$); and the number of training examples with their both views observed is $m_F=300$.}
  \label{fig:Cond2GAN_vs_MV}
\end{figure}
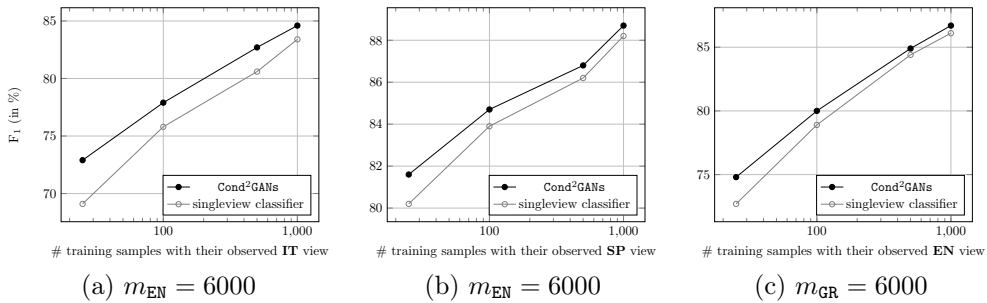

\begin{table}[h!]
  \caption{Test classification accuracy averaged over 20 random training/test sets. For each of the pairs of languages, the best result is in bold, and a $^\downarrow$ indicates a result that is statistically significantly worse than the best, according to a Wilcoxon rank sum test with $p<.01$.}
  \label{tab:AccGenviews-1}
  \centering
  \begin{tabular}{l l cc c cc c cc c cc}
    \toprule
    \cmidrule(r){1-2}
\multirow{2}{*}{Approaches} & & \multicolumn{2}{c}{(\EN,{\small $v=~$}\FR)} & & \multicolumn{2}{c}{(\EN,$v=~$\GR)} & & \multicolumn{2}{c}{(\EN,$v=~$\IT)} & & \multicolumn{2}{c}{(\EN,$v=~$\SP)}\\
        \cline{3-4}\cline{6-7}\cline{9-10}\cline{12-13}
 & & \ToneGtwo & \TtwoGone & & \ToneGtwo & \TtwoGone & & \ToneGtwo & \TtwoGone & & \ToneGtwo & \TtwoGone   \\

\cline{1-1}\cline{1-1}\cline{3-4}\cline{6-7}\cline{9-10}\cline{12-13}
\MKL & &       75.6$^\downarrow$  & 77.3$^\downarrow$ &  &  79.4$^\downarrow$ & 79.6$^\downarrow$  &  & 78.4$^\downarrow$& 79.8$^\downarrow$ & & 81.2$^\downarrow$ & 83.5$^\downarrow$     \\
\cocla & &  81.4$^\downarrow$ &83.2$^\downarrow$&       &  84.3$^\downarrow$ & 81.6$^\downarrow$& &     82.7$^\downarrow$         & 82.5$^\downarrow$& &85.1$^\downarrow$ &   86.2$^\downarrow$ \\
\mvb & &      83.1$^\downarrow$   &84.5$^\downarrow$ &       &  85.2$^\downarrow$ & 79.9$^\downarrow$ & &         84.3$^\downarrow$     & 82.1$^\downarrow$& & 84.4$^\downarrow$&   86.2$^\downarrow$   \\
\TBiGame &   & \textbf{85.3} & \textbf{85.1} &   & \textbf{86.6} & \textbf{82.9} &   & \textbf{85.3} & \textbf{84.5} &   & \textbf{86.5} & \textbf{88.3}       \\
    \bottomrule
  \end{tabular}  
\end{table} 
\paragraph{Impact of the increasing number of observed views.} In Figure \ref{fig:Cond2GAN_vs_MV}, we compare F$_1$ measures between  {\TBiGame} and one of the single-view classifiers with an increasing number of training samples, having the view corresponding to the singleview classifier observed; while the number of training examples with the other observed view is fixed.  With an increasing number of training samples, the corresponding singleview classifier gains in performance. On the other hand, {\TBiGame{}} can leverage the lack of information from training examples having their other view observed, making that the difference of performance between these models for small number of training samples is higher.

\vspace{-5mm}
\section{Conclusion}
In this paper we presented {\TBiGame} for multiview multiclass classification where observations may have missing views. The model consists of three neural-networks implementing a three players game between a discriminator and two generators. For an observation with a missing view, the corresponding generator produces the view conditionally on the other observed one. The discriminator is trained to recognize observations with a generated view from others having their views complete and to classify the latter into one of the existing classes. We evaluate the effectiveness of our approach on another 
challenging application than image analysis which is the domain of choice for the design of new GAN models. Our experiments on a subset of Reuters RCV1/RCV2 show the effectiveness of {\TBiGame} to generate high quality views allowing to achieve significantly better results, compared to the case where the missing views are generated by Machine Translation which requires a large collection of sentences and their translations to be tuned. As future study, we will be working on the generalization of the proposed model to more than 2 views. One possible direction is the use of an aggregation function of available views as a condition to the generators. 
%
%
%
%
\bibliographystyle{splncs04}
\bibliography{BibFile}

\end{document}